\pdfoutput=1
\documentclass[sigconf]{aamas}

\usepackage{balance} 
\usepackage{graphicx} 
\usepackage{algorithm}
\usepackage{algorithmic}\usepackage{latexsym}
\usepackage{amsmath}
\usepackage{amsthm}
\usepackage{float} 
\usepackage{booktabs}
\usepackage{enumitem}
\usepackage{color}
\usepackage{algorithm}
\usepackage{algorithmic}
\usepackage{bbm}
\usepackage{soul}
\usepackage{fancyhdr}
\usepackage{placeins}
\usepackage{url}
\usepackage{colortbl}
\usepackage[utf8]{inputenc}
\usepackage{makecell}
\usepackage{adjustbox}
\usepackage{tcolorbox}
\usepackage{mathtools}
\usepackage{multirow}
\usepackage{natbib}
\usepackage{multicol}
\usepackage{scontents}
\usepackage{lipsum}
\usepackage{tabularx}
\usepackage{cleveref}
\usepackage{array}
\newcolumntype{L}[1]{>{\raggedright\arraybackslash}p{#1}}

\theoremstyle{definition}

\usepackage{listings} 
\usepackage[textsize=tiny, textwidth=1.05in, colorinlistoftodos]{todonotes}

\setcopyright{ifaamas}
\acmConference[AAMAS '25]{Proc.\@ of the 24th International Conference
on Autonomous Agents and Multiagent Systems (AAMAS 2025)}{May 19 -- 23, 2025}
{Detroit, Michigan, USA}{Y.~Vorobeychik, S.~Das, A.~Nowe (eds.)}
\copyrightyear{2025}
\acmYear{2025}
\acmDOI{}
\acmPrice{}
\acmISBN{}

\acmSubmissionID{<<OpenReview submission id>>}

\title[AAMAS-2025 Formatting Instructions]
{Combining LLMs with Logic-Based Framework to Explain MCTS}
\subtitle{Extended Abstract}

\author{Ziyan An}
\affiliation{
  \institution{Vanderbilt University}\country{Nashville, US}
  }

\author{Xia Wang}
\affiliation{
  \institution{Vanderbilt University}
  \country{Nashville, US}
  }

\author{Hendrik Baier}
\affiliation{
  \institution{Eindhoven University of Technology}
  \country{Eindhoven, Netherlands}
  }

\author{Zirong Chen}
\affiliation{
  \institution{Vanderbilt University}
  \country{Nashville, US}
  }

\author{Abhishek Dubey}
\affiliation{
  \institution{Vanderbilt University}
  \country{Nashville, US}
  }

\author{Taylor T. Johnson}
\affiliation{
  \institution{Vanderbilt University}
  \country{Nashville, US}
  }

\author{Jonathan Sprinkle}
\affiliation{
  \institution{Vanderbilt University}
  \country{Nashville, US}
  }

\author{Ayan Mukhopadhyay}
\affiliation{
  \institution{Vanderbilt University}
  \country{Nashville, US}
  }

\author{Meiyi Ma}
\affiliation{
  \institution{Vanderbilt University}
  \country{Nashville, US}
  }

\begin{abstract}
In response to the lack of trust in Artificial Intelligence (AI) for sequential planning, we design a Computational Tree Logic-guided large language model (LLM)-based natural language explanation framework designed for the Monte Carlo Tree Search (MCTS) algorithm. MCTS is often considered challenging to interpret due to the complexity of its search trees, but our framework is flexible enough to handle a wide range of free-form post-hoc queries and knowledge-based inquiries centered around MCTS and the Markov Decision Process (MDP) of the application domain. By transforming user queries into logic and variable statements, our framework ensures that the evidence obtained from the search tree remains factually consistent with the underlying environmental dynamics and any constraints in the actual stochastic control process. We evaluate the framework rigorously through quantitative assessments, where it demonstrates strong performance in terms of accuracy and factual consistency. 
\end{abstract}

\keywords{Sequential Planning, Explainable AI, Large Language Model, MCTS}

\newcommand{\BibTeX}{\rm B\kern-.05em{\sc i\kern-.025em b}\kern-.08em\TeX}

\begin{document}


\pagestyle{fancy}
\fancyhead{}


\maketitle 


\begin{figure}[t]
\centering
  \includegraphics[width=0.8\linewidth]{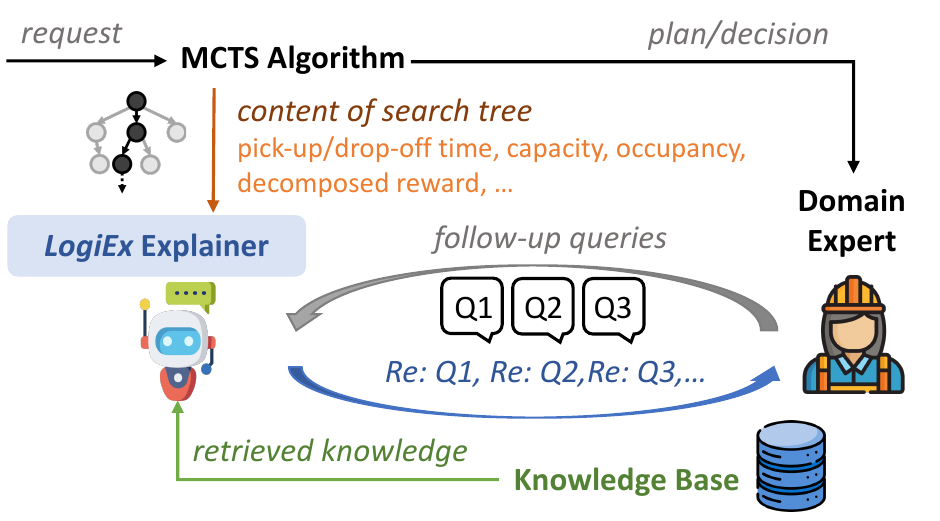}
  \caption{We explain sequential planning by combining domain knowledge, search process, and logical reasoning.}
  \label{fig:overview}
  \vspace{-1.5em}
\end{figure}

\section{Introduction}

Artificial Intelligence (AI) algorithms often operate as black-box systems, offering little to no insight into the reasoning behind their outputs. As a result, domain experts hesitate to deploy these algorithms in real-world settings due to concerns over transparency, understandability, and accountability, leaving them without a clear understanding of the implications or rationale behind the decisions made by these AI models~\cite{ma2020stlnet, ma2021toward, lundberg2017unified,ribeiro2016should,baier2021towards,marques2022delivering}.


One family of such AI approaches that is widely used in complex sequential planning problems such as manufacture engineering~\cite{saqlain2023monte} and transit route planning~\cite{weng2020pareto} is Monte Carlo Tree Search (MCTS)~\cite{kocsis2006bandit}. Understanding the results and decisions of MCTS is challenging even for experts due to the large, sampling-based search trees from which they are derived~\cite{baier2020explainable,an2024enabling}. Therefore, we develop a logic-enabled large language models (LLMs) framework that integrates knowledge and symbolic reasoning with natural language, creating a robust yet expressive xAI system for explaining planning algorithms like MCTS (Figure~\ref{fig:overview}). 

Aiming to address a flexible range of free-form user queries, our framework leverages advanced LLMs, which enables the development of xAI systems based on natural language~\cite{guha2024legalbench,chen2024sim911,welleck2021naturalproofs}. 
More specifically, it offers broad flexibility in handling queries by converting natural language inquiries submitted via a chat interface into parameterized variables and logic expressions. It then evaluates the search tree based on the criteria specified by these logic expressions, and the results are presented in the final explanation, once again expressed in natural language. The framework also enables an unlimited number of follow-up queries, facilitating an interactive, back-and-forth communication with the user.

\section{Method}

\paragraph{Background}
As the testbed for our framework, we use a paratransit planning scenario formulated as a Markov Decision Process (MDP). We define the state space, action space, constraints, and reward of the MDP. State transitions are driven by a simulated demand model for paratransit trip requests. We leverage MCTS to generate vehicle assignment decisions, which is initiated at each ``decision epoch''~\cite{joe2020deep}.

\paragraph{Query Categories and Types}
The first category of queries, called \textit{post-hoc queries}, seeks explanations for the returned plan after the algorithm has completed its execution and focuses on explaining specific MCTS decisions. The second category, called \textit{background knowledge-based queries}, focuses on the MCTS decision-making process in general.
After the user submits a query, and a Query-Classification LLM component interprets the new query and attempts to classify its intent to one of two categories. 
User queries are not restricted in terms of content or narration. However, to strategically address these queries, we pre-define 26 specific query types based on the user's underlying intentions for the first category. In contrast, for the second category, queries answerable with background knowledge, there are no specific types, as one piece of knowledge can address multiple queries. 

\paragraph{Logic Generator and Parser}
Each pre-defined query type is associated with a few-shot prompt, containing example pairs of input queries and output logic. After a new query is classified into a specific query type, the corresponding prompt is used to guide the logic generation LLM component in formulating a logic statement for the query. We categorize all user questions based on the type of evidence required to answer them: those that can be addressed with base-level evidence, referring to information directly extracted from a tree node; those that rely on derived evidence, requiring consideration of multiple nodes across different depths or branches; and those that require logic comparison evidence, involving both multi-level calculations and comparisons between two branches using Computation Tree Logic (CTL)~\cite{clarke1981design}. The variables are organized into a three-level hierarchical structure, where each level builds upon the variables and logic defined in the previous level. 

\paragraph{Logic Scorer} 
To obtain both quantitative and qualitative evidence, we define scorer functions that take the MCTS tree including states and actions as input and return either numerical or boolean values based on the evaluation of specific criteria~\cite{an2024formal}. For base-level variables, the result is obtained by identifying the target node corresponding to the variable through tree traversal. For derived evidence variables, we further define formulas to calculate the overall averaged quantitative result across all relevant nodes in the search tree. Lastly, we utilize CTL model checking algorithms to obtain logic comparison evidence, where the input is the MCTS tree. 

\paragraph{Knowledge Retrieval}
To provide domain knowledge-informed explanations for category two queries, we prepared a lightweight knowledge base containing approximately 3,000 words, divided into 34 chunks. This knowledge base covers background information on paratransit services and the MCTS algorithm, as well as detailed components of the MDP, including predefined constraints, algorithm objectives, and reward functions. We leverage the RAG technique with the OpenAI \texttt{text-embedding-3-small} model to obtain the top \( k \) results, passing information chunks to the LLM only if their relatedness scores exceed a predefined threshold.

\paragraph{Generating Explanations}
Once the list of calculated evidence or retrieved domain knowledge is obtained, the framework engages with a Question-Answering LLM to generate the final response. Key pieces of information provided to the LLM include the original user query, the evidence variables used, the result from the scorer function obtained in the previous step, and the retrieved knowledge.

\begin{table}[t]
    \centering
    \small
    \caption{Quantitative evaluation results.}
    \resizebox{\linewidth}{!}{%
    \begin{tabular}{l|l|l|l}
    \toprule
    \textbf{Method} & \textbf{Metric} & \textbf{$@$1}$\uparrow$ & \textbf{$@$3}$\uparrow$ \\
    \midrule
    Llama3.1 & FactCC / BERT & 25.77\% / 06.15\% & 34.62\% / 12.31\% \\
    Ours (Llama) & FactCC / BERT & 67.88\% / 86.54\% & \textbf{83.27\%} / \textbf{97.50\%} \\
    GPT-4o & FactCC / BERT & 42.31\% / 40.00\% & 51.15\% / 55.77\% \\
    Ours (GPT) & FactCC / BERT & \textbf{72.12\%} / \textbf{88.46\%} & 81.35\% / 94.81\% \\
    \bottomrule
    \end{tabular}
    }
    \label{tab:result-main}
    \vspace{-1.5em}
\end{table}

\section{Evaluations}
We quantitatively evaluate the framework to answer the research question (RQ): Does our framework outperform existing LLMs in generating factually accurate and relevant explanations? We consider three LLM models for our evaluation: GPT-4~\cite{achiam2023gpt}, GPT-4o~\cite{achiam2023gpt}, and Llama3.1~\cite{touvron2023llama} model. We manually prepare 620 distinct queries as inputs, each repeated three times. Accompanying each query, we manually prepare its corresponding category ID, the correct evidence variables and logic, and a reference narrative paragraph. We compare the generated explanations using two metrics: BERTScore~\cite{zhang2019bertscore} and FactCC~\cite{kryscinski2019evaluating}.

\paragraph{Factual Consistency Results and Discussions}
As shown in Table~\ref{tab:result-main}, the best result achieved across basic LLMs was a 51.15\% FactCC score, and the highest BERTScore achieved was 55.77\%. Both results suggest that basic LLMs struggle to generate relevant and factually accurate explanations directly. We then compared them with our framework with GPT-4 and Llama3.1 as backbone models, where we observed significant improvements. our framework consistently outperformed the basic LLMs across all categories. Specifically, we observed a 2.40$\times$ improvement using Llama3.1 and a 1.59$\times$ improvement using the GPT-4 model for FactCC score. The improvement in BERTScore was even more evident, with an overall increase of 7.92$\times$ for the Llama3.1 backbone model and 1.70$\times$ for the GPT-4 backbone model, respectively.

\section{Conclusion}
We present an explainability framework for MCTS sequential planning. Tested within the context of paratransit planning scenarios, our framework can address a variety of user queries by offering post-hoc explanations (after the search) and RAG-based explanations, through three-level hierarchical evidence. We thoroughly evaluated the framework performance quantitatively. Results show that our framework achieved overall superior performance compared to traditional LLMs. 


\section*{Acknowledgments}
This material is based upon work supported by the National Science Foundation (NSF) under Award Numbers 2028001, 2220401, 2151500, CNS-2238815 and CNS-1952011, AFOSR under FA9550-23-1-0135, DARPA under FA8750-23-C-0518, NWO under NWA.1389.20.251, and Horizon Europe under 101120406. The paper reflects only the authors’ view and does not necessarily reflect the views of the sponsoring agencies.



\bibliographystyle{ACM-Reference-Format} 
\bibliography{sample}


\end{document}